\documentclass{IEEEcsmag}

\usepackage[colorlinks,urlcolor=blue,linkcolor=blue,citecolor=blue]{hyperref}
\expandafter\def\expandafter\UrlBreaks\expandafter{\UrlBreaks\do\/\do\*\do\-\do\~\do\'\do\"\do\-}
\usepackage{upmath,color}
\usepackage{amsfonts}
\usepackage[caption=false,font=normalsize,labelfont=sf,textfont=sf]{subfig}
\usepackage{stfloats}
\usepackage{subcaption}

\jvol{XX}
\jnum{XX}
\paper{8}
\jmonth{Month}
\jname{Publication Name}
\jtitle{Publication Title}
\pubyear{2024}

\DeclareRobustCommand*{\IEEEauthorrefmark}[1]{%
  \raisebox{0pt}[0pt][0pt]{\textsuperscript{\footnotesize\ensuremath{#1}}}}

\makeatletter
\def\@IEEEpubid{}%
\def\@IEEEpubidadjcol{}%
\makeatother
\pubid{} % clear IEEE copyright block

\begin{document}

\sptitle{Regular Paper}

\title{Using Kolmogorov-Smirnov Distance for Measuring Distribution Shift in Machine Learning}

\author{
    \IEEEauthorblockN{O. K. Tonguz \IEEEauthorrefmark{1}, F. Taschin \IEEEauthorrefmark{2}}\\
    \IEEEauthorblockA{\IEEEauthorrefmark{1}Carnegie Mellon University, USA}
    \IEEEauthorblockA{\IEEEauthorrefmark{2}KTH Royal Institute of Technology, Sweden}
}
\markboth{THEME/FEATURE/DEPARTMENT}{THEME/FEATURE/DEPARTMENT}

\begin{abstract} 

One of the major problems in Machine Learning (ML) and Artificial Intelligence (AI) is the fact that the probability distribution of the test data in the real world could deviate substantially from the probability distribution of the training data set. When this happens, the predictions of an ML system or an AI agent could involve large errors which is very troublesome and undesirable. While this is a well-known hard problem plaguing the AI and ML systems' accuracy and reliability, in certain applications such errors could be critical for safety and reliability of AI and ML systems. One approach to deal with this problem is to monitor and measure the deviation in the probability distribution of the test data in real time and to compensate for this deviation. In this paper, we propose and explore the use of Kolmogorov-Smirnov (KS) Test for measuring the distribution shift and we show how the KS distance can be used to quantify the distribution shift and its impact on an AI agent's performance. Our results suggest that KS distance could be used as a valuable statistical tool for monitoring and measuring the distribution shift. More specifically, it is shown that even a distance of KS=0.02 could lead to about 50\% increase in the travel time at a single intersection using a Reinforcement Learning agent which is quite significant. It is hoped that the use of KS Test and KS distance in AI-based smart transportation could be an important step forward for gauging the performance degradation of an AI agent in real time and this, in turn,  could help the AI agent to cope with the distribution shift in a more informed manner.

\end{abstract}

\maketitle

The efficient control of vehicle flow to reduce travel times and CO2 emissions has become critically important as efforts for sustainable urban mobility intensify. Consequently, research in Traffic Signal Control has gained significant traction in recent years, as smarter and more efficient traffic lights can mitigate congestion, decrease travel times, lower emissions, and enhance road safety. In this domain, Deep Reinforcement Learning (DRL) has garnered considerable attention from the research community due to its ability to effectively address the sequential decision-making nature of traffic management problems. DRL algorithms can generate optimal Traffic Signal Control policies by dynamically responding to observed traffic patterns through interactions with a realistic traffic simulator in a trial-and-error manner.

However, DRL agents are susceptible to the Distribution Shift problem, which occurs when the probability distribution of the test data deviates from that of the training data. In the context of Traffic Signal Control, this issue arises when the traffic patterns observed during testing differ from those during training. Such variations can result from natural changes in human behavior throughout the day, alterations in city infrastructure, new traffic regulations, or unforeseen events such as accidents or roadworks. In these instances, the performance of the machine learning (ML) system or an AI agent can deteriorate significantly, since the Traffic Signal Control policy learned during training may no longer be optimal. Although Distribution Shift is a major challenge in deploying DRL algorithms in real-world scenarios, the existing literature has not yet offered a comprehensive framework to measure it. Developing a framework to assess Distributional Shift in Traffic Signal Control scenarios, which can predict a DRL agent's performance degradation, is crucial not only for evaluating the performance of DRL agents but also for detecting potential performance risks during deployment and taking preemptive actions.

In this study, we propose to measure Distribution Shift using the Kolmogorov-Smirnov (KS) distance between the training and test distributions, characterized by the proportion of vehicle flow across different traffic volumes. We present empirical results from both real-world and synthetic data, demonstrating the impact of Distribution Shift on the performance of DRL agents in Traffic Signal Control scenarios.

\section{RELATED WORK}

Research on the Distribution Shift problem spans several decades across information theory and statistics, with seminal contributions such as \cite{pac} standing out for its foundational influence. Over the years, various studies have underscored this issue, spanning from classical models not rooted in Deep Learning, as observed in \cite{concept_drift} and \cite{bayram2022concept}, to more contemporary approaches like \cite{wiles2021finegrained}, which delve into correcting for distribution shifts. Notably, advancements in deep object recognition have been marked by innovations such as the Domain-Adaptive Network architecture \cite{domain_adaptive} and techniques in unsupervised adaptation \cite{unsupervised_adaptation}. Interestingly, even fundamental components like the BatchNorm layer \cite{batchnorm} were originally conceived to mitigate distribution shift challenges in Deep Models. These techniques have significantly bolstered the robustness of Deep Learning models, although they often lack a systematic means to quantify the impact of distributional disparities on model performance.

In recent years, the application of Reinforcement Learning (RL) to optimize Traffic Signal Control has attracted considerable interest, promising optimal policy learning for traffic management scenarios. Pioneering works like \cite{2016rl}\cite{dqn_tsc} pioneered the use of DQN RL algorithms \cite{mnih2013playing} with handcrafted reward functions in this domain. Recent advancements have demonstrated the efficacy of Deep RL in diverse settings, including large-scale deployments \cite{presslight} and scenarios involving partial observability \cite{tonguz2021}, yet these studies typically assume negligible distribution shift between training and testing phases, an assumption often untenable and unrealistic in practical applications.

Efforts to measure the impact of Distribution Shift have been documented extensively in Machine Learning literature \cite{dataset_shift}, extending into specific domains such as Language Modeling \cite{djolonga2021robustness} and Image Classification \cite{recht2019imagenet} within Deep Learning. However, the focus on Distributional Shift in Reinforcement Learning, particularly in the context of Traffic Signal Control, remains underexplored. While \cite{generalight} proposed a method for measuring statistical distances between traffic distributions, primarily for training generative models in traffic simulation, its adaptation for evaluating the performance impact of Distribution Shift on ML systems remains limited. Notably, \cite{taori2020measuring} offers a broader framework for assessing Distribution Shift effects during model testing, although it does not specifically address Traffic Signal Control scenarios.

In our research, we aim to bridge this gap by focusing on proactive measurement of Distribution Shift effects before model deployment. Our objective is to predict the critical thresholds of Distributional Shift where significant performance degradation in Traffic Signal Control models becomes evident, thereby enhancing the robustness and reliability of such systems in real-world applications.

\section{TRAFFIC SIGNAL CONTROL AND REINFORCEMENT LEARNING}

\subsection{Traffic Signal Control}
\label{sec:traffic_signal_control}
\begin{figure}
    \centering
    \includegraphics[width=0.45\textwidth]{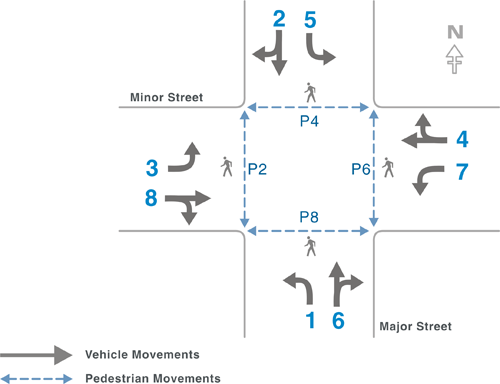}
    \caption{NEMA phases for a standard 4-way intersection. Each arrow represents a traffic
    movement, and some phases serve multiple movements.}
    \label{fig:nema_phases}
\end{figure}

Traffic Signal Control is the problem of deciding when and how to switch the traffic lights at each
signalized intersection in a road network to serve incoming vehicles efficiently. An intersection
is a point where two or more roads meet, and it is signalized if it has traffic lights regulating
the flow of vehicles. Each road branch connected to the intersection is called an
\textit{approach}, and each approach can have one or more lanes. Each lane is associated with a
\textit{movement}. In the US, traffic movements are grouped in \textit{NEMA phases} (or just
phases), which, for a standard 4-leg intersection, are 8 in total (see Figure
\ref{fig:nema_phases}).

A Traffic Signal Control Policy is an algorithm that decides which phase to serve at any time.
Simpler policies can be based on fixed-time schedules (which is still prevalent in many cities and countries), while more advanced policies can be based on real-time observations of the traffic at the intersection obtained from sensors such as cameras, radars, or
Lidars.

\subsection{Reinforcement Learning}
\label{sec:reinforcement_learning}
The Traffic Signal Control problem is formulated in Reinforcement Learning as a Markov Decision
Process \cite{bellman1957markovian}, with state space $S$, action space $A$, and a reward function
$r: S \times A \times S \rightarrow \mathbb{R}$. A state is an observation of the traffic at the
intersection (e.g., queue length for each phase) and an action is a decision on which phases to
serve. The reward is a scalar value based on "how good" the current state is. Transitions between states are stochastic and are regulated by the -unknown- transition probability function $P(s' | s, a)$ that relates the probability of a new state $s'$ given the current state $s$ and the chosen action $a$. In other words, at any time step, the agent observes the current state $s$, selects an action $a$, receives a reward $r$, and transitions to a new state $s'$ according to the transition probability function $P(s' | s, a)$.

The goal is to discover an optimal policy $\pi: S \rightarrow A$ that maps the observed state of
the intersection to the optimal action such that the sum of all future rewards from that
state-action pair is maximized. In value-based Deep Reinforcement Learning, a Neural Network is
trained to learn a function $Q^*: S \times A \rightarrow \mathbb{R}$ that maps each state-action pair to the expected sum of future
returns, discounted by a factor $\gamma$:
\begin{align}
    \label{eq:optimal_q}
    Q^*(s, a) = \max_{\pi}\mathbb{E}_{\pi^*}\left[\sum_{k=t}^{\infty} \gamma^{t-k} r_k | s_t = s, a_t = a\right]
\end{align}

All experiments in this paper will use a value-based Deep Reinforcement Learning algorithm called
DQN \cite{mnih2013playing} which learns the optimal $Q$ function by interacting with a
simulated environment and training the Neural Network to minimize the Bellman Squared Loss function below
with Stochastic Gradient Descent \cite{ruder2016overview}.
\begin{align}
\label{eq:bellman_loss}
    \mathcal{L}(s, a, r, s') = \left(Q(s, a) - r(s, a) + \gamma \max_{a'}Q(s', a')\right)^2
\end{align}
where $s'$ is the state we end up in by choosing action $a$ in state $s$. Repeatedly minimizing this
loss from transitions $(s, a, r, s')$ collected by interacting with the environment generally
produces a Neural Network that estimates with good approximation the optimal $Q^*$ function and
allows one to choose optimal actions that maximize $Q$ in any state $s$.

\section{Kolmogorov-Smirnov Test}
The Kolmogorov-Smirnov (K-S) test is a non-parametric test of the equality of two continuous,
one-dimensional probability distributions that can be used to compare a sample distribution with
a reference probability distribution. The test is non-parametric because it does not require an
analytical form for the data distribution, and can be performed between a known reference
distribution and a sample, or between two sample distributions. The K-S test is based on the
empirical Cumulative Distribution Function (CDF) of the two distributions under comparison. Given
 a probability distribution $f(x)$, its CDF $F(x)$ is defined as the probability that a random
 variable $X$ takes on a value less than or equal to $x$, as shown in Equation \ref{eq:cdf}.

\begin{equation*}
  F_X(x) = P(X \le x) = \int_{-\infty}^{x} f(x) dx
  \label{eq:cdf}
\end{equation*}

\begin{figure}
  \centering
  \includegraphics[width=0.45\textwidth]{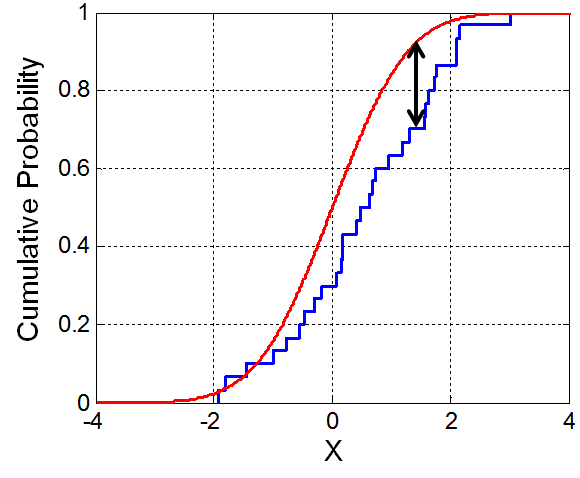}
  \caption{K-S distance $D$ (black line) between a reference distribution (red) and an empirical
  distribution (blue).}
  \label{fig:ks_plot}
\end{figure}

Central to the K-S test is the K-S statistic (or distance) $D$ which represents the maximum
difference between the Cumulative Distribution Function (CDF) of the two distributions being
compared. Given the CDFs $F_1(x)$ and $F_2(x)$ of the two distributions, the K-S statistic is
defined as shown in Equation \ref{eq:ks_statistic}. The K-S distance $D$ is visually represented
by the maximum vertical distance between the two CDFs, as shown in Figure \ref{fig:ks_plot}.

\begin{equation}
  D = \sup_{x} |F_1(x) - F_2(x)|
  \label{eq:ks_statistic}
\end{equation}

The K-S test is based on the null hypothesis that the two distributions are identical. If the
two distributions are identical, the K-S distance $D$ converges to 0 as the number of samples $n$
goes to infinity. Generally, for a value of $n$ that approaches infinity, the distribution of
$\sqrt{n}D$ converges to the Kolmogorov distribution, which is independent of the reference
distribution \cite{ks_test}. This result, also known as Kolmogorov Theorem, allows us to reject
the null hypothesis with confidence $\alpha$ if
\begin{equation*}
    D > K_{\alpha, n}
\end{equation*}
where $K_{\alpha, n}$ is the critical value of the K-S statistic at significance level $\alpha$ 
for sample size $n$, which is pre-computed from the Kolmogorov distribution and available in the 
Kolmogorov-Smirnov table.

\section{Measuring Distribution Shift in Traffic Scenarios}
\label{sec:measuring_ds}
We propose to quantify the discrepancy between two traffic scenarios using the Kolmogorov-Smirnov
(KS) statistic to compare the normalized distribution of vehicles across the 8 NEMA phases in the
two scenarios. Considering an intersection as depicted in Figure \ref{fig:nema_phases} that
incorporates all 8 phases, we define a scenario as a specific time period (e.g., 1 hour)
characterized by a particular vehicle flow for each available traffic movement. This definition
aligns with common practices in the literature for train/test scenarios and is also measurable by
advanced Traffic Signal Monitoring systems such as Automated Traffic Signal Performance Measures
(ATSPM) \cite{atspm}, from which we obtain our real-world data.

\paragraph{Traffic Distribution}
We define the traffic distribution of a scenario as the normalized distribution of vehicles
across the 8 NEMA phases. Given a 1-hour scenario with $n$ total vehicles, we represent it as a
8-categorical distribution with the 8 NEMA phases as support and the normalized number of vehicles
in each phase as the corresponding probability. Given $N_i$ as the number of vehicles in each phase
$i$, the traffic distribution is therefore defined as

\begin{equation}
  p(i) = \frac{N_i}{n}
  \label{eq:traffic_distribution}
\end{equation}

\paragraph{K-S Distance in Traffic Distributions}
We use the K-S distance of Equation \ref{eq:ks_statistic} differently from a traditional K-S
test. First of all, we are not interested in answering a yes-no question such as "Are the two
distributions identical?" but rather in quantifying the discrepancy between the two distributions.
Secondly, the results on the significance of rejecting the null hypothesis that we discussed in
the previous Section are valid for continuous distributions, while we characterize traffic as
a discrete distribution as in Equation \ref{eq:traffic_distribution}. The K-S test can
be extended to discrete distributions \cite{ks_discrete} keeping the same formula for the
K-S distance. For these reasons, we use the K-S distance as a measure of discrepancy between
two traffic distributions. Given two traffic distributions $p_A$ and $p_B$, the K-S distance
is defined as shown in Equation \ref{eq:ks_traffic}.

\begin{equation}
  D = \max_{i} |p_A(i) - p_B(i)|
  \label{eq:ks_traffic}
\end{equation}

By using the K-S distance as a measure of discrepancy between two traffic distributions, we are
implicitly assuming that the maximum difference between the two CDFs is the most important
metric to quantify the discrepancy between the two traffic distributions. We believe this
assumption is reasonable, since the effect of small difference in the traffic volume of a single
phase might be negligible in the overall performance, while large differences can have a great 
impact. 

\begin{figure*}
    \centering
    \includegraphics[width=\linewidth]{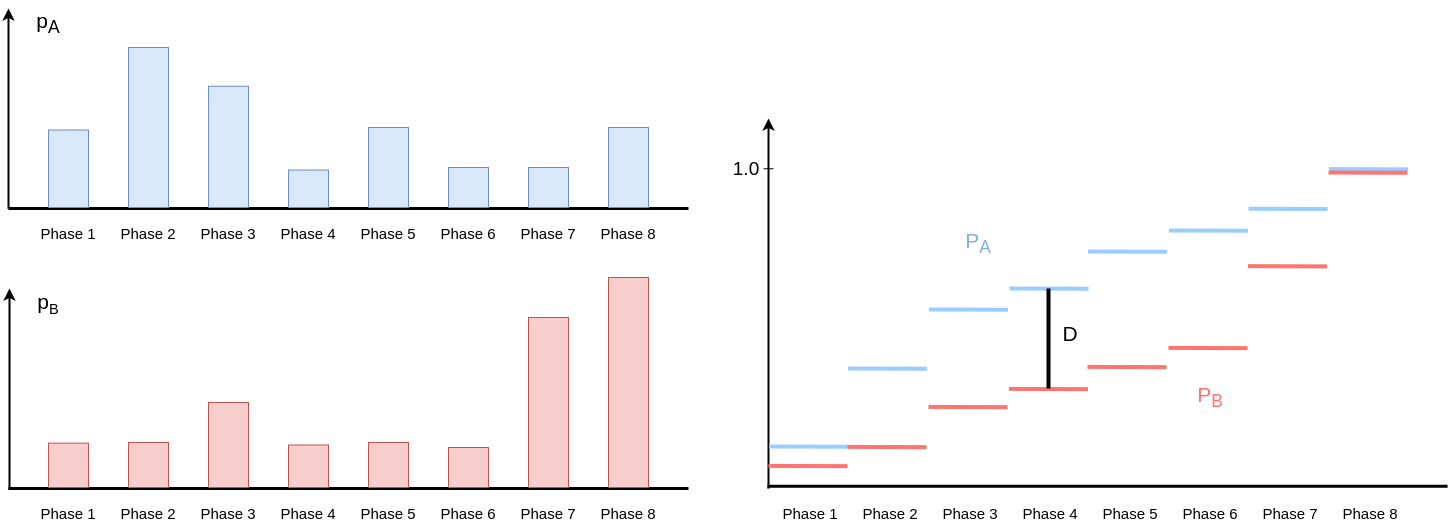}
    \caption{On the left, the two traffic distributions $p_A$ (blue) and $p_B$ (red). On the right, their corresponding CDFs and the K-S distance highlighted in black.}
    \label{fig:enter-label}
\end{figure*}

 % ---------------------------------------------------------
\section{Experiments}
\label{sec:experiments}
In this Section, we present the experimental approach followed in all our experiments. First, we describe
how we generate scenarios from real-world data, how we train and test the DRL agent, and the performance metrics we measure. Then, we present all the experiments.

\subsection{Simulated Scenarios}
\label{sec:simulated_scenarios}
The real-world data we base our experiments on is collected from the ATSPM system, which is capable
of providing turn-count data for a large number of intersections in the state of Utah, in United States (US). ATSPM provides
vehicle volume data for each traffic movement in time buckets of 5, 15, and 60 minutes. We use the 60-minute data to generate scenarios for our experiments. In particular, for all our experiments we use a 4-way intersection in Orem, Utah, between State Street and Center Street. To  simulate traffic scenarios, we use our own simulator which is based on the SUMO engine. We use our software pipeline to generate the SUMO map file of the intersection from Open Street Maps (OSM) \cite{osm} data and the SUMO route files from the ATSPM data. Route files contain each vehicle  in the simulation along with its origin, destination, and scheduled departure time. Given the  values for vehicle flow for each traffic movement in the 60-minute period, we generate the route files by randomly assigning a departure time to each vehicle in the given period of time with a
 uniform distribution. Our simulator advances the simulation in steps of 1 second and asks for an action from the agent at every step. Actions only represent the combination of phases to serve and do not include the yellow-red transitions which are managed by our simulator. At every step, the agent is able to choose only among a set of valid actions (e.g., it cannot set green to a phase that is undergoing a yellow-red transition) so that we ensure the
 correctness of the simulation and the safety of the traffic management system.

\subsection{Agent Training}
\label{sec:agent_training}
As discussed in Section \textit{Traffic Signal Control and Reinforcement Learning}, we use a DQN agent to learn the optimal
Traffic Signal Control policy. We allow the agent to observe the number of incoming vehicles for each possible phase in a range of 30m (this is based on the range of the radars used as sensors at the aforementioned intersection in Orem, Utah, between State Street and Center Street), corresponding to the actual detection range of sensors at the intersection into consideration. We augment the observation space with the current color of each phase and the elapsed time in that color. The agent can choose among 8 possible actions, one for each phase pair, and is rewarded based on the number of vehicles that cross the intersection at every time step.

\subsection{Performance Measures}
\label{sec:performance_measures}
To measure the DRL agent performances we use two metrics: the Normalized Throughput and the Extended Travel Time (the details of the latter metric is explained in the Appendix).

\paragraph{Normalized Throughput} The normalized throughput is defined as the ratio between the number of vehicles that crossed the intersection in the given period of time and the total number of vehicles that were supposed to be generated in that period.

\paragraph{Extended Travel Time} The Extended Travel Time is the time that passes between the moment a vehicle is scheduled to depart and the moment it arrives to its destination. This includes any delay that the vehicle might experience, including a delay in its departure if the departure
point is occupied by other vehicles at that time. Note that the departure delay can be non-negligible in high-traffic scenarios where the intersection is often congested.

\subsection{Experiment 1: Distribution Shift in Real World Scenarios}
\label{sec:exp_real_world_scenarios}
This experiment serves as an initial demonstration of the existence of the Distribution Shift
problem in Traffic Signal Control. In this experiment, we generate 4 scenarios from real-world
data at 4 different 1-hour periods (7am-8am, 9am-10am, 2pm-3pm, 5pm-6pm) during the same day
(Tuesday March 14th 2023). The training set corresponds to the 1-hour period between 7am and 8am,
which contains 3978 vehicles  for which we generate 10 training scenarios by only shuffling the
departure times of the vehicles to ensure a more varied training set. We train our agent as described in Section \textit{Agent Training} above and test it on all 4 scenarios to evaluate the level of performance degradation due to the distribution shift.

\subsection{Experiment 2: Total vehicle volume vs phase KS distance}
\label{sec:exp_total_vehicle_volume_vs_phase_KS_distance}
As discussed in the \textit{Measuring Distribution Shift in
Traffic Scenarios} Section, it is not possible to attribute the performance degradation of the agent to only distributional shift without considering the total vehicle volume as well. This experiment serves as a first attempt to decouple the two phenomena (KS distance and
total vehicle volume), by keeping one constant and varying the other. We start from the
training scenario of the previous \textit{Experiment 1},
i.e., the 1-hour period between 7am and 8am. From it, we derive two sets of scenarios:
fixed-volume and fixed-phase-distribution. The fixed-volume set contains 11 scenarios where we
synthetically modify the distribution of vehicles across the 8 phases for increasing values of
phase KS distance from 0 to 1.0 while maintaining the total vehicle volume intact. The fixed-phase-distribution set instead contains 7 scenarios where we synthetically vary the total vehicle volume (from 2000 to 7000) while maintaining the phase distribution intact. We train our DRL agent on the original scenario and test it on all these scenarios and we observe the performance degradation. 

\subsection{Experiment 3: large scale study}
\label{sec:exp_all_total_vehicle_volume_and_phase_KS_distance}
In this experiment, we generate synthetic scenarios starting from the same training scenario of
\textit{Experiment 1} (7am-8am) in the same way as we did in \textit{Experiment 2}, but this time we vary both the total vehicle volume and the phase KS distance with large granularity. First, we select 7 different phase distributions with increasing values of phase KS distance from 0.0 (the training scenario) to 0.6 and 13 values of total vehicle volume from 4000 to 7000 with steps of 250. Observe that a phase distribution is a vector with 8 values that sum to 1 and describes how vehicles are distributed across the 8 phases, irrespective of the total volume. Then, we generate the corresponding scenarios for each combination of the two, for a total of 91 scenarios. We evaluate the agent trained on the training scenario in each of these test scenarios and plot the results in x-y plots with total vehicle volume on the x-axis and the agent performance (throughput and extended travel time) on the y-axis, color-coded by phase KS distance. This allows us to observe two things: a) how the
overall performance changes as the total vehicle volume increases, and b) how the phase KS distance affects the agent's performance.

% ---------------------------------------------------------
\section{Results}
\label{sec:results}
Here we report the results for the three experiments mentioned above
maintaining their name and order for ease of reference.

\subsection{Experiment 1: Distribution Shift in Real World Scenarios}
\label{sec:res_real_world_results}

Table \ref{tab:real_world_travel_time} shows the throughput ratio for the 4 real-world scenarios of \textit{Experiment 1} described in the \textit{Experiments} section above. We observe that the throughput ratio decreases as the phase KS distance increases. Although this confirms that the phase KS distance is predictive of DRL agent performance degradation, it is not enough to determine the effect of this degradation as the scenarios have different total vehicle volumes. We analyze the impact of the increasing volume and KS distance in the two following experiments.

%\begin{figure}
%    \centering
%    \includegraphics[width=0.45\textwidth]{images/real_world_throughput.png}
%    \caption{Throughput ratio for the 4 real-world scenarios. The x axis represents the phase KS
%    distance of each scenario from the training one (the data point with phase KS distance 0.0).}
%    \label{fig:real_world_throughput}
%\end{figure}
%\begin{figure}
%    \centering
%    \includegraphics[width=0.45\textwidth]{images/real_world_ext_travel_time.png}
%    \caption{Extended Travel time for the 4 real-world scenarios. The x-axis represents the phase KS
%    distance of each scenario from the training one (the data point with phase KS distance 0.0).}
%    \label{fig:real_world_ext_travel_time}
%\end{figure}

\begin{table*}
\centering
\begin{tabular}{|c|c|c|c|}
\hline
Scenario & KS Distance & Normalized Throughput & Extended Travel Time [s] \\
\hline
7-8 am & 0 & 0.98 & 74.29  \\
9-10 am & 0.032 & 0.89 & 220.55  \\
2-3 pm & 0.067 & 0.70 & 315.48  \\
5-6 pm & 0.069 & 0.66 & 635.39 \\
\hline
\end{tabular}
\vspace{7pt}
\caption{Comparison between Travel Time and Extended Travel Time metrics for the 4 real-world
scenarios. In the second column, we report the KS Distance of each scenario to the training scenario}
\label{tab:real_world_travel_time}
\end{table*}

\subsection{Experiment 2: Total vehicle volume vs phase KS distance}
\label{sec:res_total_vehicle_volume_vs_phase_KS_distance}
\begin{figure*}
\centering
\subfloat[Fixed total vehicles volume]{\includegraphics[width=0.45\textwidth]{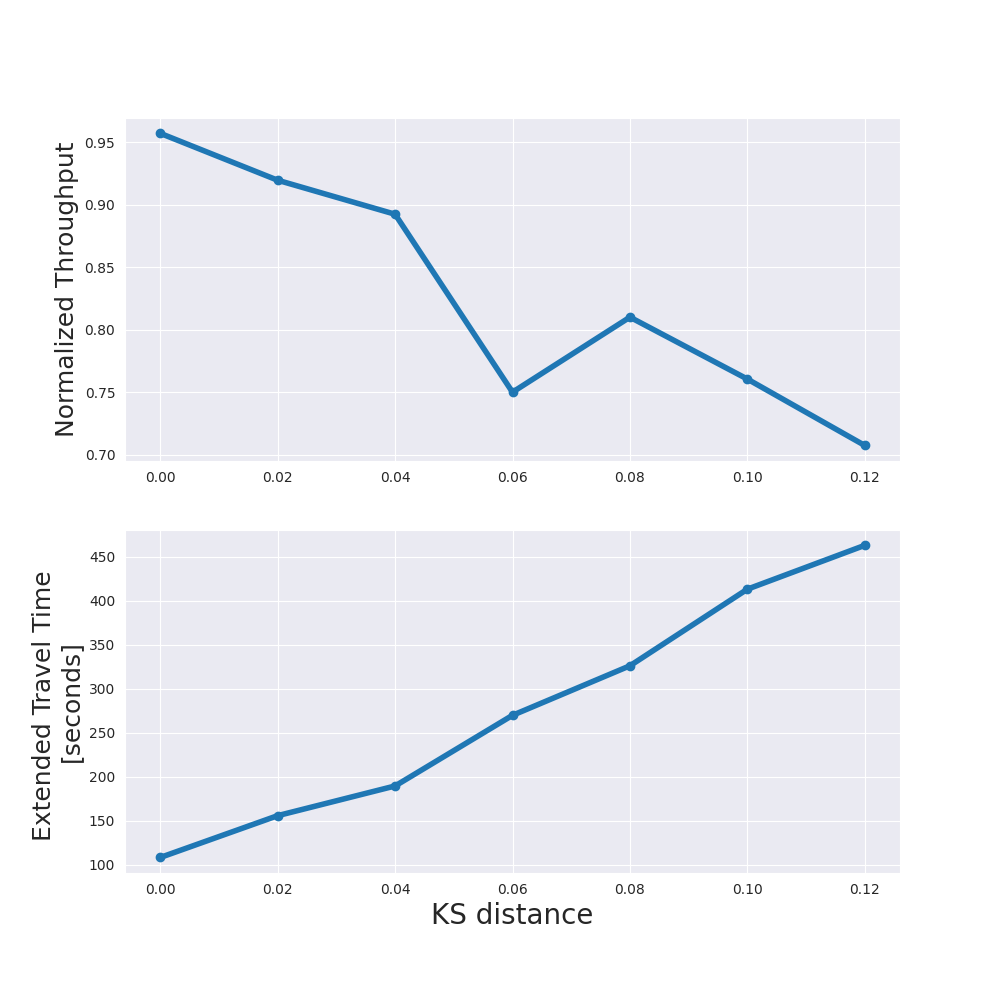}}
\subfloat[Fixed phase distribution]{\includegraphics[width=0.45\textwidth]{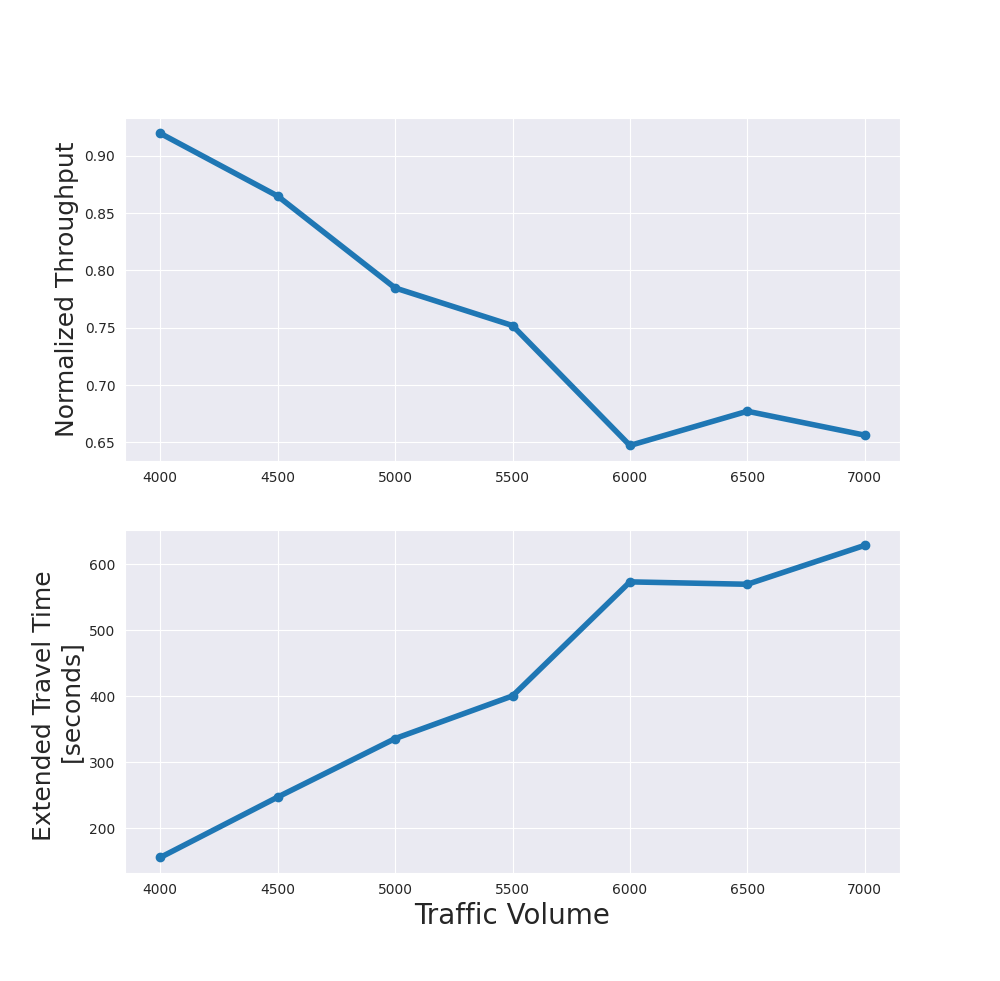}}
\caption{Throughput ratio and Extended Travel Time for the fixed total vehicles volume and fixed phase distribution scenarios. In Figure (a), the x-axis represents the phase KS distance from the training scenario. In Figure (b), the x-axis represents the total volume of vehicles.}
\label{fig:fixed_volume_vs_fixed_phase}
\end{figure*}

Figure \ref{fig:fixed_volume_vs_fixed_phase} shows the results for the fixed-volume and
fixed-phase-distribution cases of \textit{Experiment 2}.
Regarding the fixed-volume case, Figure \ref{fig:fixed_volume_vs_fixed_phase}(a), we observe that
the throughput ratio decreases as the phase KS distance increases, which is consistent with our
hypothesis that the phase distribution's KS distance is predictive of DRL agent performance degradation.
Similarly, the Extended Travel Time is affected the phase KS distance. In
the fixed-phase-distribution case, Figure \ref{fig:fixed_volume_vs_fixed_phase}(b), we observe
a U-shaped behavior, where the throughput ratio is higher for the 4k vehicles case -which is almost identical to the training scenario- and decreases in both directions of increasing and decreasing total vehicle volume. The same behavior is observed for the Extended Travel Time.

\subsection{Experiment 3: All combinations of total vehicle volume and phase KS distance}
\label{sec:res_all_total_vehicle_volume_and_phase_KS_distance}
\begin{figure*}
\centering
\subfloat[Normalized Throughput]{\includegraphics[width=0.50\textwidth]{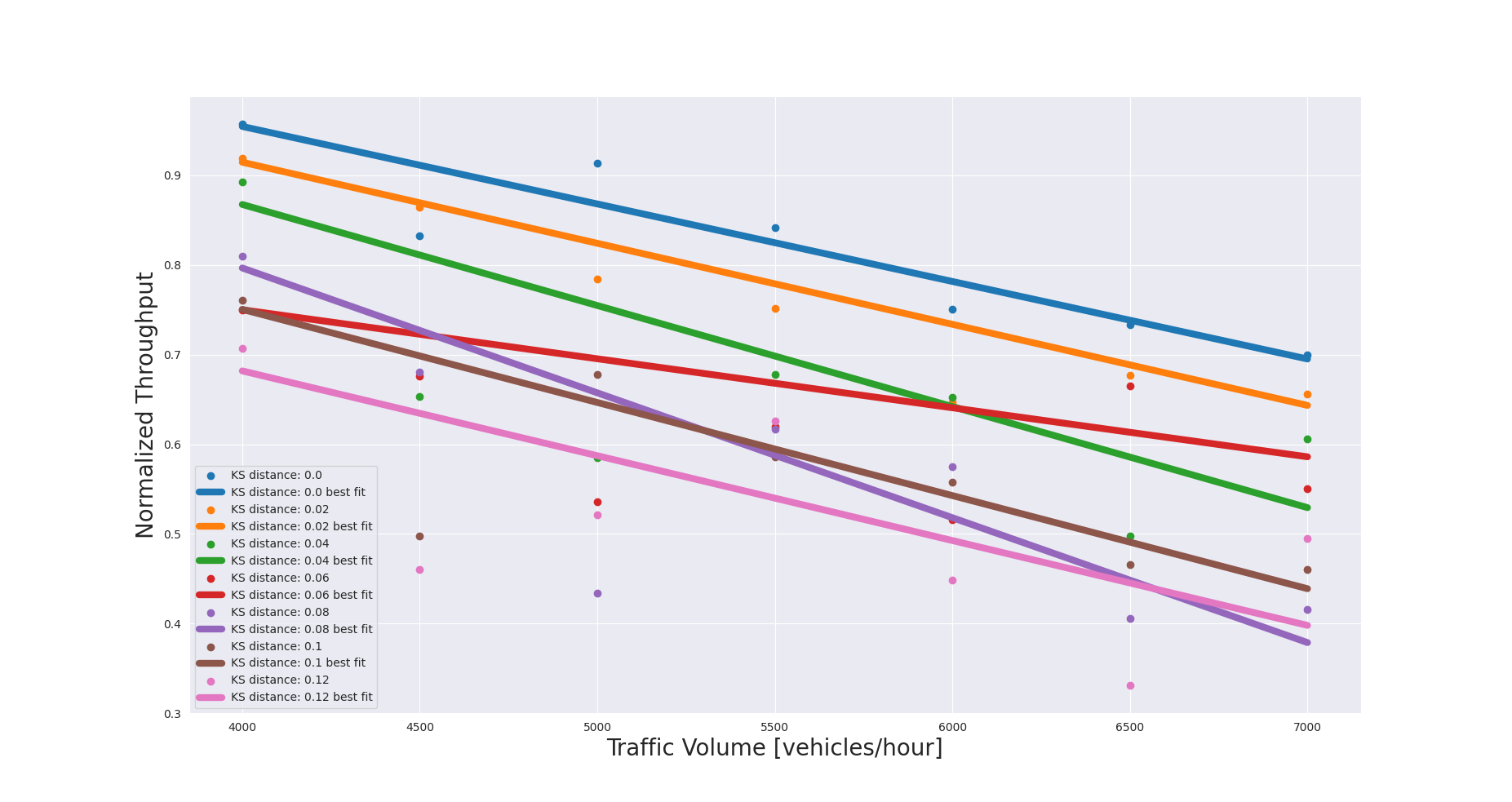}}
\subfloat[Extended Travel Time]{\includegraphics[width=0.50\textwidth]{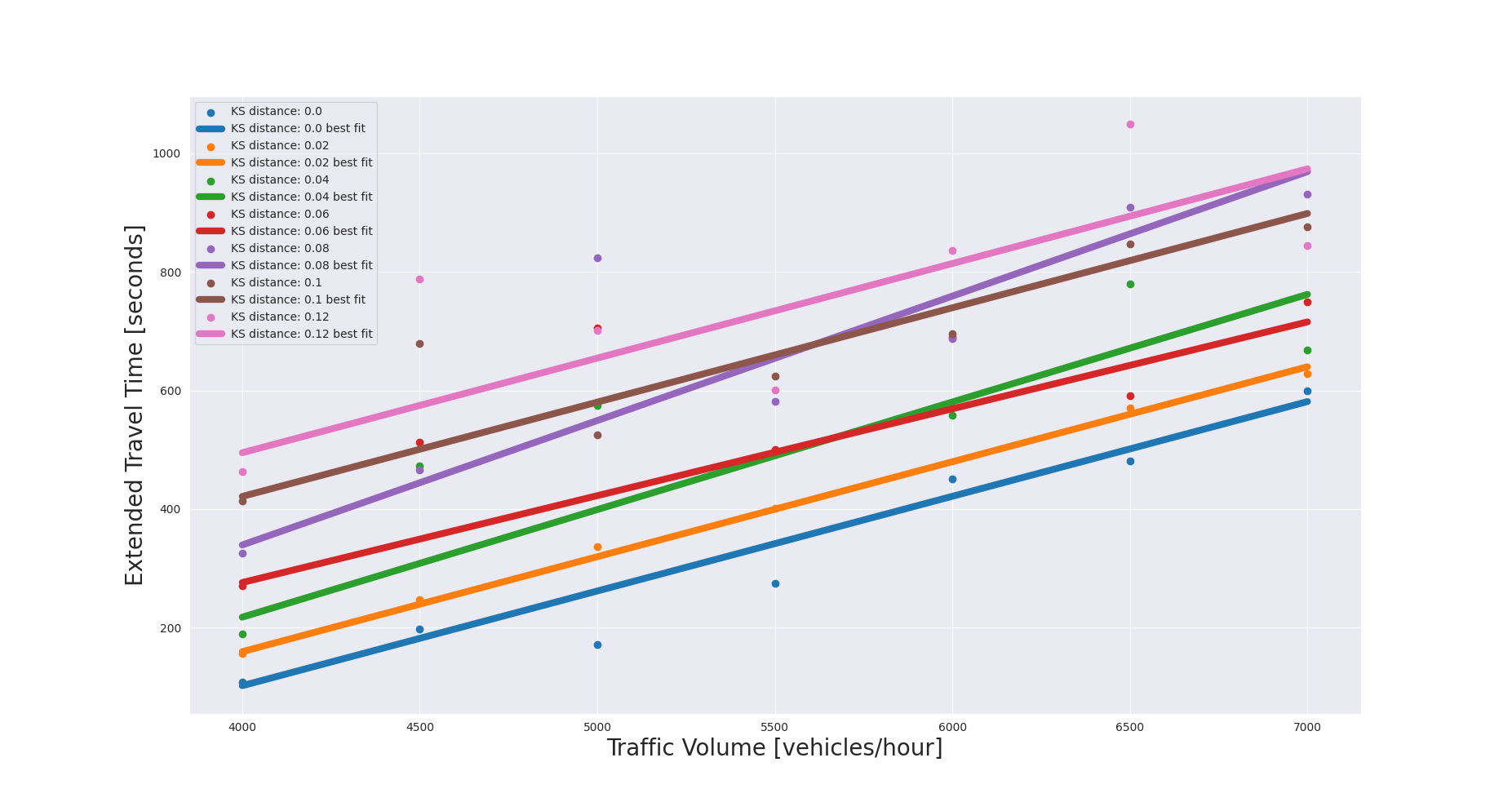}}
\caption{Throughput ratio and Extended Travel Time for all combinations of total vehicle volume and
phase KS distance. The x axis represents the total vehicle volume, the y axis represents the agent
performance, and the different curves are colored according to their KS distance (blue = 0.0, orange = 0.1, green = 0.2, red = 0.3, purple = 0.4, brown = 0.5, pink = 0.6)}
\label{fig:all_total_vehicle_volume_and_phase_KS_distance}
\end{figure*}

Figure \ref{fig:all_total_vehicle_volume_and_phase_KS_distance} shows the results for all
combinations of total vehicle volume and phase KS distance described in \textit{Experiment 3} of Section \textit{Experiments}. In Figure
\ref{fig:all_total_vehicle_volume_and_phase_KS_distance}(a), we can observe several curves of
throughput ratio vs total vehicle volume, color-coded by phase KS distance and the best-fit
linear approximation for each. As expected, an increase in total volume of vehicles leads to a decrease in throughput ratio. The phase KS distance, however, has a significant
impact on the curves, where curves corresponding to lower phase KS distances have a higher
throughput ratio for the same total vehicle volume. In Figure
\ref{fig:all_total_vehicle_volume_and_phase_KS_distance}(b), we observe that the
Extended Travel Time also increases with the total volume of vehicles and the impact of increasing KS distances is significant.

\section{Discussion}
\label{sec:discussion}
Our results show that when utilizing a Reinforcement Learning (RL) agent at an intersection with real-world data, the distribution shift problem manifests itself in conjunction with changes in the volume of vehicles arriving at that intersection. This phenomenon complicates isolating the impact of distribution shift alone. To assess the impact of distribution shift resulting from changes in vehicle volume (including the number of vehicles in each of the 8 NEMA phases), we examined the effects of these factors on the agent's performance both separately and jointly. This approach enables us to accurately determine and quantify the impact of distribution shift on the agent's performance.

The significant impact of distribution shift is clearly demonstrated in the results of Experiment 2, where it is separated from the effect of increased vehicle volume. As established in Queuing Theory and Operations Research, an increase in load or demand in a queuing system typically leads to increased delays, as illustrated in Figure 3(b). Results from Experiment 3 not only highlight how distribution shift qualitatively affects the performance of a DRL agent but also allow us to quantify this effect. For example, an increase in KS Distance of 0.02 results in an average loss of 3.7\% in Normalized Throughput, while an increase of 500 vehicles leads to a 4.3\% loss. Additionally, travel time increases by 42\% for a 0.02 increase in KS Distance, compared to a 76\% increase for an additional 500 vehicles per hour, underscoring the pronounced effect of distribution shift. These results show that distribution shift has a substantial impact on performance. The significant rise in travel time, even with a fixed distribution, illustrates the sensitivity of DRL agents to both distribution shift and overall vehicle volume increases.

To put this in context, consider the following: if a Department of Transportation intends to automate the use of an RL agent deployed at an intersection, it could set a threshold KS Distance of 0.04, at which point the travel time roughly doubles, as a criterion for discontinuing the use of the trained DRL agent.

\paragraph{Why Use the KS Distance?}
Using the KS (Kolmogorov-Smirnov) distance could be very helpful because it captures the largest deviation in the cumulative distribution function between two distributions. This is particularly useful in traffic signal control scenarios, where small deviations in the distribution across each of the 8 phases might be inconsequential, but a significant deviation in any single phase can have a substantial impact. The authors of this paper observed this phenomenon on US-89 that runs parallel to I-15 in the state of Utah.

While other statistical measures such as the Wasserstein distance consider the overall mismatch between distributions, the KS distance focuses on the largest discrepancy between the two distributions, making it more sensitive to significant deviations in specific phases. This characteristic is critical in scenarios such as corridor management, where substantial deviations might occur in the main corridor phases, significantly affecting traffic flow. In contrast, deviations in phases on lateral approaches might be negligible. Therefore, the KS distance is particularly suitable for identifying and responding to impactful distribution shifts that could adversely affect the performance of a DRL agent in traffic signal control.

Despite the linearity of our results (i.e., the relationship between an increase in KS distance and the agent's performance is linear), in Appendix B we show that the relationship between KS distance and the cumulative difference between train and test distributions is not linear. This means that using the KS distance as a measure allows us to obtain linearity in performance results that would not hold if we were using, for example, the cumulative difference as a statistical distance measure. 

\section{Conclusion}
\label{sec:conclusion}
In this paper, we propose a method to measure the Distribution Shift in Traffic Signal Control scenarios based on the Kolmogorov-Smirnov (KS) divergence and we show how an increase in this measure can adversely affect the performance of DRL agents in Traffic Signal Control scenarios. We present quantitative results on the relationship between an increase in KS divergence and the corresponding degradation in an AI agent's performance. To the best of our knowledge, this is the first work that specifically addresses the problem of how to measure Distributional Shift in Traffic Signal Control scenarios using KS distance. As future work, it'd be interesting to compare these results with results obtained using other statistical measures of distribution shift (e.g., Wasserstein) and to include other parameters in the computation, such as the arrival time of vehicles.

%\printbibliography

\section{Appendices}

\subsection{Appendix A: Non-linearity of KS distance}
\begin{figure}
    \centering
    \includegraphics[width=\linewidth]{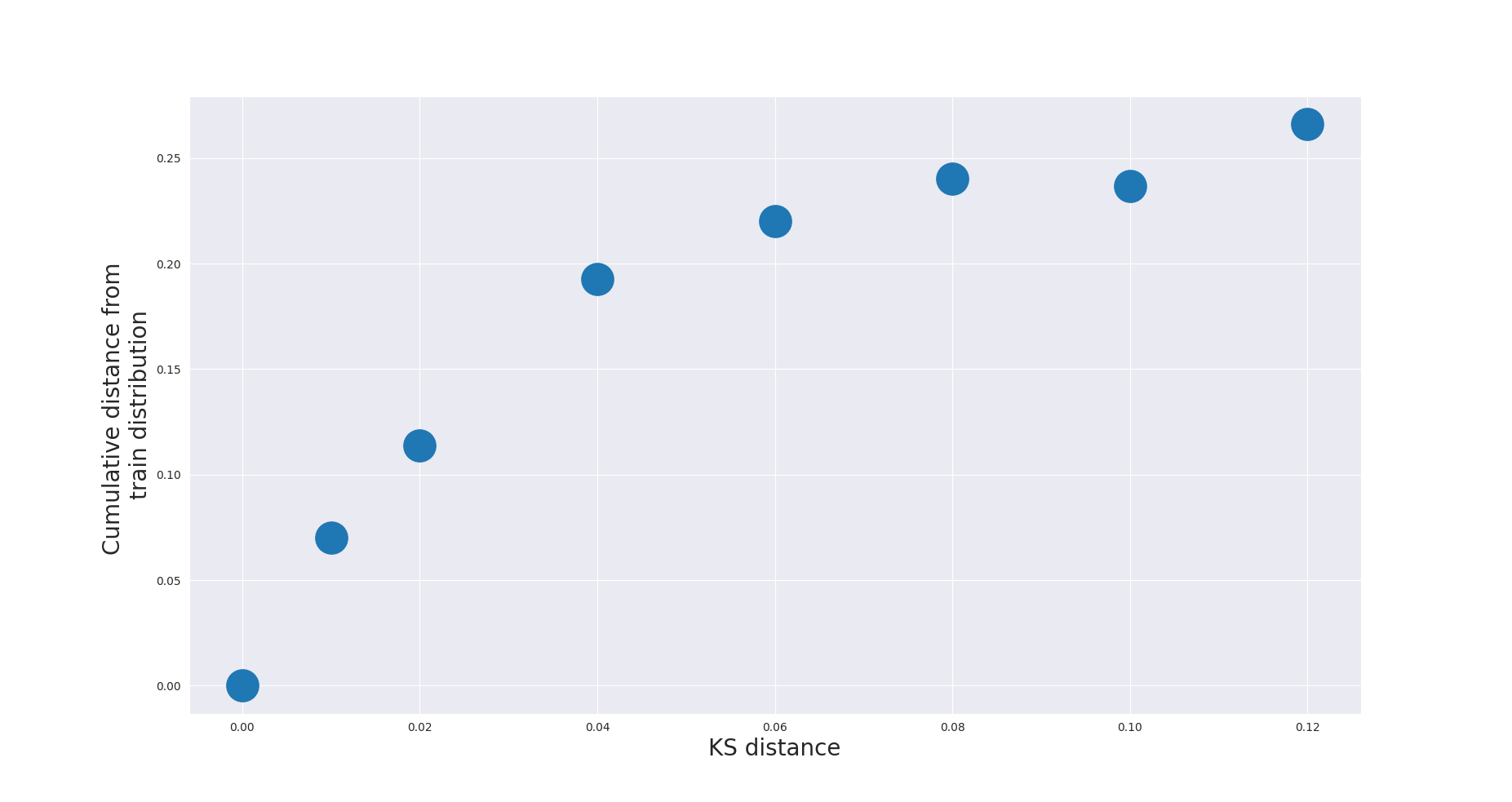}
    \caption{Absolute difference between the train distribution and the test distributions from increasing values of the KS distance. Observe that 50\% of the maximum deviation is obtained for KS = 0.02, while all other steps to KS = 0.12 lead to the remaining 50\%.}
    \label{fig:ks_dist_cumulative}
\end{figure}

In the Results section, we observe how small increases in KS distance can have a large impact on throughput and travel time. Surprisingly, even the first step (KS = 0.02) already leads to a significant degradation in performances. This leads us to two questions: a) How much does the distribution change with an increase in KS distance? b) Is every increase in KS equal? To address these questions, we compute the \textit{cumulative difference} from the training distribution that corresponds to every KS value in Experiments 2 and 3 of the Results Section. Given the training distribution $P_{train}$ and a test distribution $P_{KS=x}$ with KS distance equal to $x$, we compute the cumulative difference $D_{cumulative}$ for a KS distance level $x$ as
\begin{equation}
    D_{cumulative}(x) = \sum_{i}|P_{train}(i) - P{KS=x}(i)|
\end{equation}
where $P_{train}(i)$ and $P{KS=x}(i)$ are the proportion of vehicles in phase $i$ for the training and test distribution, respectively. Figure \ref{fig:ks_dist_cumulative} shows how the cumulative difference from the train distribution increases as a function of KS distance. From the figure, we can clearly observe how the first steps in KS distance are those that lead to a larger increase in cumulative difference (i.e., that makes the test distribution deviate from the training distribution). Interestingly, even at KS = 0.02, the test distribution is 12\% different from the train distribution, while at KS = 0.12 (6 times more)  the difference only doubles. This clearly shows that even small values of KS distance implies significant deviations from the training distribution.

\subsection{Appendix B: Extended Travel Time Breakdown}
\begin{figure}[h!]
    \centering
    \includegraphics[width=\linewidth]{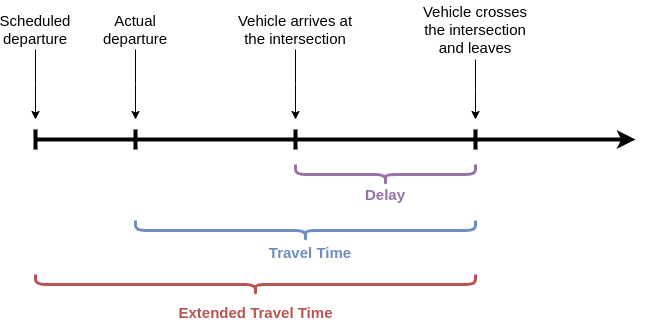}
    \caption{Comparison of Extended Travel Time, Travel Time, and Delay metrics. The Extended Travel Time includes all components of delay in a vehicle's trip, while Travel Time excludes the departure delay. The Delay metric (obtained by subtracting the free flow time from the purple portion) considers only vehicles at the intersection approaches. }
    \label{fig:time_metrics}
\end{figure}

In this Appendix, we examine the Extended Travel Time metric in detail to explain how we measure it and discuss its significance. A traffic scenario in our simulation environment is defined by a file containing each vehicle's path and departure time. During the simulation, the actual departure time of vehicles can be later than the expected one, as their departure road might be occupied by other vehicles. This departure delay is an important portion of the total vehicle's delay which should not be neglected. This is especially the case for high volume of traffic or congested intersections. Therefore, instead of using the Travel Time metric; i.e., the time between departure and arrival, we use the Extended Travel Time metric, where we consider the scheduled departure time instead of the actual departure time. In Figure \ref{fig:time_metrics} we show a breakdown of the Extended Travel Time metric and we compare it with simple Travel Time and Delay, which are often used in the literature. Travel Time is the same, but it excludes the departure delay. The Delay measure typically used in the literature measures the time from when the vehicle enters one of the intersection approaches to when it leaves, and is obtained by subtracting the free-flow time from this value. 

\begin{figure}[h!]
\centering{\includegraphics[width=\linewidth]{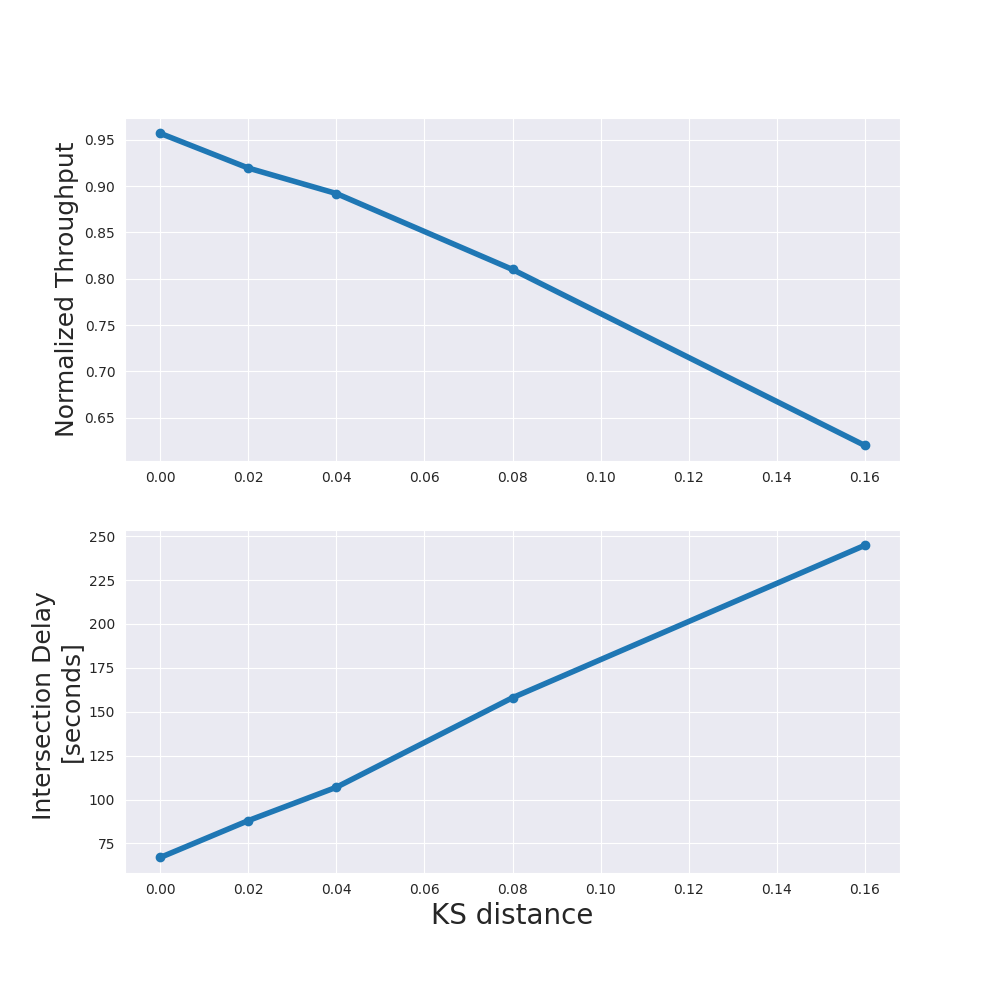}}
\caption{Intersection Delay for the fixed total vehicles volume scenarios, for increasing values of KS distance of 0.0, 0.02, 0.04, 0.08, 0.16}
\label{fig:fixed_volume_delay}
\end{figure}

In Figure \ref{fig:fixed_volume_delay} we show the results of Experiment 2 in terms of Delay as a function of the KS distance. It can be observed that the values on the y-axis are substantially lower than those in all other plots. This highlights the difference between the Intersection Delay metric and the Extended Travel Time metric, where the latter captures much more of a vehicle's journey at an intersection. 

%\begin{figure}
%   \centering
%   \includegraphics[width=\linewidth]{images/intersection_delay.png}
%    \caption{Throughput ratio and Intersection Delay for all combinations of total vehicle volume and phase KS distance. The x-axis represents the total vehicle volume, the y-axis represents the agent performance, and the different curves are colored according to their KS distance (blue = 0.0, orange = % 0.02, green = 0.04, red = 0.06, purple = 0.08, brown = 0.1, pink = 0.12)}
%    \label{fig:all_total_vehicle_volume_and_phase_KS_distance_delay}
%\end{figure}

\end{document}